\documentclass[]{interact}
\usepackage{float}

\usepackage{epstopdf}
\usepackage[caption=false]{subfig}
\usepackage{etex}
\usepackage[etex=true,export]{adjustbox}
\usepackage{multirow}
\usepackage{hyperref}

\usepackage[numbers,sort&compress]{natbib}
\bibpunct[, ]{[}{]}{,}{n}{,}{,}
\renewcommand\bibfont{\fontsize{10}{12}\selectfont}
\makeatletter
\def\NAT@def@citea{\def\@citea{\NAT@separator}}
\makeatother

\theoremstyle{plain}
\newtheorem{theorem}{Theorem}[section]

\theoremstyle{definition}

\theoremstyle{remark}

\def\E{{\mathbb E}}
\def\V{{\mathbb V}}

\begin{document}

\articletype{}

\title{Expertise-based Weighting for Regression Models with Noisy Labels}

\author{
\name{M.R. Santos\textsuperscript{a,b}\thanks{CONTACT M.R. Santos Email: milene.santos@usp.br} and Rafael Izbicki\textsuperscript{b}}
\affil{\textsuperscript{a}University of São Paulo, BR; \textsuperscript{b}Federal University of São Carlos, BR}
}

\maketitle

\begin{abstract}
Regression methods assume that accurate labels are available for training. However, in certain scenarios, obtaining accurate labels may not be feasible, and relying on multiple specialists with differing opinions becomes necessary.  Existing approaches addressing noisy labels often impose restrictive assumptions on the regression function. In contrast, this paper presents a novel, more flexible approach. Our method consists of two steps: estimating each labeler's expertise and combining their opinions using learned weights. We then regress the weighted average against the input features to build the prediction model. The proposed method is formally justified and empirically demonstrated to outperform existing techniques on simulated and real data. Furthermore, its flexibility enables the utilization of any machine learning technique in both steps. In summary, this method offers a simple, fast, and effective solution for training regression models with noisy labels derived from diverse expert opinions.
\end{abstract}


\section{Introdution}

Supervised learning regression methods aims to find a function $g(\textbf{x})$ that predicts a real label $\textbf{Y} \in \Re$ based on the input covariates $\textbf{X} = (X_1,X_2, \ldots, X_d)$. 
In order to do that, it is often assumed one
has access to a labeled dataset, $(\textbf{X}_1,Y_1),\ldots,(\mathbf{x}_n,Y_n)$.
However, in many situations, obtaining the real labels $Y_i$'s may be costly or time-consuming, or even impossible to collect. In such scenarios, asking different experts to provide their opinions regarding the real label for each observation is a common practice. 
As experts  only provide an educated guess of the true value of $Y_i$, it is common to ask many experts to provide their estimate of such quantity. 
Cases like these include spam detection, diagnosis of patients based on images and morphological
classification of galaxies \citep{J,LI,RI,SC,F}.
We denote such opinions by $Y_{i,1},\ldots,Y_{i,J}$, where $J$
is the number of experts.
A key question is how to best use this information to train $g$. 

The standard approach to deal with such noisy labels is to use the average of the opinions, $J^{-1}\sum_{j=1}^J Y_{i,j}$, as a proxy for the true value $Y_i$, and then use standard supervised learning methods
using this proxy as the label to be predicted \cite{X,GU,CI}. It is known however that such an approach is suboptimal, especially in settings where the experts have different expertise (that is, some are more accurate than others) \citep{Y}. Thus, much work has been done on alternative ways to use such information.

Most approaches however only deal with classification problems, that is, problems where $Y$ is quantitative \citep{R,Y,C,I,ZHE,T,Z}. In a regression context,
existing approaches deal only with specific settings. For instance,
\citep{R} develops an iterative method based on the Expectation-Maximization algorithm assuming a linear relationship between the true label, $Y$, and the features. \citep{RO} developed a method specific  to topic modelling that also uses a parametric assumption.  \citep{G} and \citep{X} propose methods based on Gaussian Processes. Thus, the literature lacks of procedures that can have good performance under more general scenarios.

In this paper, we propose a simple, fast, and powerful method to train $g$ in a regression context while taking into account the diversity between labellers. Our approach, named WEAR (Weighted Expertise-based Average Regression), has two steps: first, we estimate the expertise of each labeller by estimating its variability. Then, we create a weighted average of $Y_{i,1},\ldots,Y_{i,J}$ using the learned weights. Finally, we regress the weighted average on $\mathbf{x}$ to create our prediction model. Any machine learning method can be used in both steps, yielding flexibility to the method.

Section \ref{sec:method} presents a formal justification of the method, as well as additional details. Section \ref{sec:experiments} presents empirical evidence on both simulated and real data that our approach has better performance than competing approaches. Finally, Section \ref{sec:final} concludes the paper.

\section{Proposed Method}
\label{sec:method}

\subsection{Motivation}

Our method begins by attempting to recover the true labels, drawing inspiration from the following theorem. The proof of this theorem can be found in the Appendix.

\begin{theorem}
\label{thm:main}
Let $Y_j$ be the opinion of the $j$-th expert,  $j=1,\ldots,J$, $Y$ be the real label and $\mathbf{x}$ be the vector of covariates.
Assume that:
\begin{itemize}
    \item $Y,Y_1,\ldots,Y_J$ are independent,
    \item For every $j=1,\ldots,J$, $\E[Y_j| \mathbf{X} ]=\E[Y|\mathbf{X}],$ that is, the experts are unbiased.
\end{itemize}
Then, the solution for   
\begin{equation}
\label{min}
 \arg \min_{w_1,\ldots,w_{J}: \sum_j w_j=1} \E\left[\left(\sum_{j=1}^J  w_j Y_j - Y\right)^2 \middle\vert \mathbf{X} = \mathbf{x} \right],
\end{equation}
is to take
$$w_i = \frac{\V^{-1}[Y_i\mid \mathbf{X} = \mathbf{x}]}{\sum_{j=1}^J \V^{-1}[Y_j\mid \mathbf{X} = \mathbf{x}]}.$$
\end{theorem}

The theorem shows that the most effective way to linearly combine experts and achieve accurate recovery of Y is by weighting them based on the reciprocal of their variance. 
To estimate these variances, we make a simplifying assumption that $\V[Y_j\mid \mathbf{X} = \mathbf{x}]$ remains constant across all values of $\mathbf{x}$. The following theorem outlines a method for estimating the optimal weights under this assumption.

\begin{theorem}
\label{thm:consistent} 
Let  $r_i(\mathbf{x}):=\E[Y_i|\mathbf{x}]$ be the regression of the label given by the $i$-th
expert on $\mathbf{x}$.
 Then, under the assumptions of Theorem \ref{thm:main} and if
$\V[Y_j\mid \mathbf{X} = \mathbf{x}]$ is constant in $\mathbf{x}$, 
a consistent estimator of the optimal weight
$w_i$
is 
\begin{align*}
    \frac{R_i^{-1}}{\sum_{j=1}^J R_j^{-1}},
\end{align*}
where
\begin{align*}
    R_i:= \frac{1}{m} \sum_{k=1}^m \left(Y'_{k,i}- r_i(\mathbf{x}'_k) \right)^2
\end{align*}
is the mean squared error of regression $r_i$
and  $\{(\mathbf{x}'_k,Y'_{k,1},\ldots,Y'_{k,J})\}_{k=1}^m$ is a validation sample.
\end{theorem}

It also follows from this theorem that $R^{-1}_i$ is a proxy for the expertise of the $i$-th expert.

In practice, $r_i(\mathbf{x})$ is also unknown and therefore must be estimated. The next section details our full procedure.

\subsection{WEAR: Weighted Expertise-based Average Regression}

Our method consists of the following steps:

\begin{enumerate}
\item Separate the data set into training, validation and test data sets. 

\item Estimate $r_j(\mathbf{x}):=\E[Y_j|\mathbf{x}]$, the regression of the  label given by the $j$-th expert on $\mathbf{x}$, using the training set. Let $\widehat r_j$ be such estimate.

\item Use the validation data set to estimate $R_i$ from Theorem \ref{thm:consistent}:
$$\widehat R_j:= \frac{1}{m} \sum_{k=1}^m \left(Y'_{k,i}- \widehat r_j(\mathbf{x}'_k) \right)^2$$
and approximate the optimal weights of Theorem \ref{thm:main}:
 $$\widehat w_i := \frac{\widehat R_i^{-1}}{\sum_{j=1}^J \widehat R_j^{-1}}.$$

\item Compute $\overline{Y}^{w}_k$, the weighted mean of the experts on the $k$-th training sample point, using 
\begin{equation*}
\overline{Y}^{w}_k= \sum_{j=1}^{J}\widehat w_j Y_{k,j}.
\end{equation*}

\item Estimate the true regression function,
$r(\mathbf{x}):=\E[Y|\mathbf{x}]$, by regressing 
$\overline{Y}^{w}_k$ on $\mathbf{x}_k$ using the training sample.

\end{enumerate}

WEAR offers great flexibility as we can utilize any machine learning algorithm in steps 2 and 4. For instance, if we expect only a few covariates to be associated with $Y$ and $Y_j$'s, we can use lasso, random forests, or sparse additive models. These algorithms perform variable selection, and studies have shown that they work well in such scenarios \citep{RAV,B}. Alternatively, if we expect $\mathbf{x}$ to lie on a submanifold of $\Re^d$, k-nearest neighbors, support vector regression, or spectral methods might be a better option. These algorithms have shown success in situations where $\mathbf{x}$ is expected to be on a submanifold of $\Re^d$ \citep{K,ST,L,I_L}.
Similary, of $\mathbf{x}$
represents image data,  our method can
easily leverage convolutional networks to estimate the regression functions.

The following section demonstrates that this method, despite its simplicity, not only delivers excellent predictive performance but also achieves a high degree of accuracy in identifying the true expertise of each expert ($w_j$).

\section{Experiments}
\label{sec:experiments}
\subsection{Simulated Data}
\label{sec:experiments_simulated}

 In order to evaluate the performance of WEAR, we used simulated data with a known gold standard, which allowed us to compare the results of our proposed model with existing methods. In all settings,
we used 10,000 observations to train the model, and 5,000 as the validation set, as described in Section \ref{sec:method}. 85,000 additional sample points were used to compare the predictive performance of the methods that were investigated through the Mean Square Error (MSE) of the true label.

To estimate the expertise and fit the proposed model, we used several methods, including linear regression, Tree, Forest, and Lasso. We also used the Raykar's Algorithm \cite{R} as a baseline. We also added regression methods that use the true labels as a baseline (which would be unavailable in practice). Concretely,
 the fitted models were:
\begin{itemize}
    \item \textbf{Our methods:} \begin{itemize}
        \item LINEAR REGRESSION WITH WEIGHTED MEAN, linear regression via least squares method using the weighted mean from this work as the dependent variable,
        \item FOREST WITH THE WEIGHTED MEAN, random forest method using the weighted mean from this work as the dependent variable,
        \item TREE WITH WEIGHTED MEAN, regression tree method using the weighted mean from this work as the dependent variable
        \item LASSO WITH WEIGHTED MEAN, lasso regression method with the weighted mean from this work as the dependent variable.
    \end{itemize}
    \item \textbf{Baselines that can be computed using real data:} 
    \begin{itemize}
       \item RAYKAR ALGORITHM, the method shown in \cite{R},
\item LINEAR REGRESSION WITH THE ARITHMETIC MEAN, linear regression via least squares method using the arithmetic mean as the dependent variable,
\item FOREST WITH ARITHMETIC MEAN, random forest method using the arithmetic mean as the dependent variable,
\item TREE WITH ARITHMETIC MEAN, regression tree method using the arithmetic mean as the dependent variable,
\item LASSO WITH ARITHMETIC MEAN, Lasso regression method using the arithmetic mean as the dependent variable,
    \end{itemize}
    \item \textbf{Baselines only available on simulated data:} 
    \begin{itemize}
\item LINEAR REGRESSION WITH REAL $Y$, linear regression via least squares method using the real label as the dependent variable,
\item FOREST WITH REAL $Y$, random forest method using the real label as the dependent variable,
\item TREE WITH REAL $Y$, regression tree method using the real label as thedependent variable, 
\item LASSO WITH REAL $Y$, Lasso regression method using the real label as the dependent variable,
    \end{itemize}
\end{itemize}

We used R software (\url{https://www.r-project.org/}) for the analysis, along with specific packages for the Forest, Tree, and Lasso methods: randomForest \citep{RC}, rpart \citep{TH}, and glmnet \citep{H}, respectively. We calculated the Lasso model hyperparameters using cross-validation, and use the default values for the tree-based methods. To avoid any bias due to simulation randomness, we generated 100 different samples with the same characteristics. The final result for the MSE presented in the tables represents the mean MSE across those 100 samples.

We investigate four settings. In two of them (1 and 2), the true regression function $r(\mathbf{x})$ is nonlinear on the covariates, and in two of them (3 and 4), it is linear.
Moreover, in two of them (1 and 3), experts have a similar expertise), while in two of them (2 and 4), they do not; some are clearly better than others. More specifically,  the settings we investigate are given by
\begin{itemize}
    \item \textbf{Experiment 1}:\\  $Y = 2x_1^2 + 1x_2 + 5x_3+0.5x_4+4x_5^2+3x_6^2 + \epsilon$;\\ $\V[Y_1|\mathbf{x}]=4$, $\V[Y_2|\mathbf{x}]=4.41$, $\V[Y_3|\mathbf{x}]= 4.84$,  $\V[Y_4|\mathbf{x}]= 5.0625$
    \item \textbf{Experiment 2}:\\  $Y = 2x_1^2 + 1x_2 + 5x_3+0.5x_4+4x_5^2+3x_6^2 + \epsilon$; \\ $\V[Y_1|\mathbf{x}]=4$, $\V[Y_2|\mathbf{x}]=100$, $\V[Y_3|\mathbf{x}]= 2500$,  $\V[Y_4|\mathbf{x}]= 10000$
    \item \textbf{Experiment 3}:\\  $Y = 2x_1 + 1x_2 + 5x_3+0.5x_4+4x_5+3x_6 + \epsilon$;\\ $\V[Y_1|\mathbf{x}]=4$, $\V[Y_2|\mathbf{x}]=4.41$, $\V[Y_3|\mathbf{x}]= 4.84$,  $\V[Y_4|\mathbf{x}]= 5.0625$
    \item \textbf{Experiment 4}: \\ $Y = 2x_1 + 1x_2 + 5x_3+0.5x_4+4x_5+3x_6 + \epsilon$; \\$\V[Y_1|\mathbf{x}]=4$, $\V[Y_2|\mathbf{x}]=100$, $\V[Y_3|\mathbf{x}]= 2500$,  $\V[Y_4|\mathbf{x}]= 10000$
\end{itemize}
We always choose $\epsilon \sim N(0,3^2)$ and each expert is generated according to $Y_i|\mathbf{x},y \sim N(y,\V[Y_i|\mathbf{x}])$, all experts independent of each other.

Table \ref{tab:performance_simulated} presents the mean squared error (MSE) for each model and experiment, along with its standard error. The best-performing models are highlighted in bold. The results of our analysis are summarized as follows:
\begin{itemize}
    \item Our proposed approach (WEAR) achieves results that are comparable to the gold standard in all experimental settings, which is not  available in real-world datasets.
    \item For the linear settings (experiments 3 and 4), our method performs similarly to \citep{R}, a model that was specifically designed for this type of problem.
    \item When the experts have comparable expertise (experiments 1 and 3), WEAR yields results that are comparable to those obtained using a simple arithmetic mean. This indicates that our approach does not sacrifice performance by adding more parameters
    \item  WEAR's flexibility enables it to outperform linear models in the nonlinear settings (experiments 1 and 2), where nonlinear models perform better. This is in contrast to \citep{R}, which assumes linear relationships.
\end{itemize}

Overall, these findings demonstrate the effectiveness of our proposed approach and its versatility in handling different experimental settings.

\begin{table}[H]
\caption{Estimated Mean Squared Error (MSE) results and corresponding standard errors (in parentheses) for real datasets. Our approach demonstrates comparable performance to methods relying on the gold standard label Y, which is typically unavailable in practical scenarios.}
\begin{adjustbox}{width=\columnwidth,center}
\label{tab:performance_simulated}
\begin{tabular}{l|l|c|c|c|c}
\hline
\textbf{Framework}                            & \textbf{Model}                            & \textbf{Experiment 1} & \textbf{Experiment 2} & \textbf{Experiment 3} & \textbf{Experiment 4} \\ \hline
  \multirow{4}{*}{
  \shortstack[l]{WEAR\\ (Our approach)}} & Linear Regression & 271.18 (1.9)          & 268.83 (2.2)          & \textbf{9.00 (0.0)}            & \textbf{9.01 (0.0) }           \\
& Random Forest            & \textbf{73.45 (2.2)}           & \textbf{70.00 (2.3)}           & 10.07 (0.0)           & 10.20 (0.0)           \\
& Regression Tree                & 79.12 (1.3)           & 77.02 (1.0)           & 17.28 (0.0)           & 17.58 (0.0)           \\
& Lasso                     & 271.19 (1.9)          & 268.83 (2.2)          & \textbf{9.00 (0.0)}            & \textbf{9.01 (0.0)}            \\ \hline
  Raykar Algorithm  & --                          & 271.18 (1.9)          & 268.83 (2.2)          & \textbf{9.00 (0.0)}            & \textbf{9.02 (0.0) }           \\ \hline
  \multirow{4}{*}{Arithmetic Mean}  & Linear Regression   & 271.18 (1.9)          & 269.39 (2.2)          & \textbf{9.00 (0.0)}            & 9.55 (0.0)            \\
& Random Forest              & \textbf{73.90 (1.3)}           & 96.43 (2.2)           & 10.07 (0.0)           & 39.43 (0.0)           \\
& Regression Tree                & 78.92 (1.3)           & 98.86 (3.0)           & 17.28 (0.0)           & 27.10 (0.0)           \\
&Lasso                 & 271.19 (1.9)          & 269.39 (2.2)          & \textbf{9.00 (0.0)}            & 9.55 (0.0)            \\ \hline
  \multirow{4}{*}{
  \shortstack[l]{
  Real $Y$\\ (Gold standard)}} & Linear Regression              & 271.17 (1.9)          & 268.83 (2.2)          & \textbf{9.00 (0.0)}            & \textbf{9.01 (0.0)  }          \\
& Random Forest                       & \textbf{73.15 (2.1)}           & \textbf{69.51 (2.4)}           & 10.03 (0.0)           & 10.03 (0.0)           \\
& Regression Tree                          & 78.06 (1.3)           & 76.65 (0.9)           & 17.25 (0.0)           & 17.25 (0.0)           \\
& Lasso                       & 271.19 (1.9)          & 268.84 (2.2)          & \textbf{9.00 (0.0)}            & \textbf{9.01 (0.0) }           \\ \hline
\end{tabular}
\end{adjustbox}
\end{table}

The precision in the estimation of each expert's variance is evaluated across different datasets and methods, and the results are summarized in Table \ref{tab:weights}. The table presents the average absolute deviation between each expert's estimated variance and its true value. In linear settings (experiments 3 and 4), \citep{R} provides reliable variance estimates. However, in nonlinear settings (experiments 1 and 2), WEAR outperforms other methods and yields more accurate estimates of expert precision.

\begin{table}[H]
\caption{Average absolute deviation between each expert's estimated variance and its true value  on the simulated datasets. WEAR provides a more accurate estimate of the experts' precision in nonlinear settings.}
\label{tab:weights}
\begin{adjustbox}{width=\columnwidth,center}
\begin{tabular}{l|l|c|c|c|c}
\hline
\textbf{Framework} & \textbf{Model}                      & \textbf{Weight Experiment 1} & \textbf{Weight Experiment 2} & \multicolumn{1}{l|}{\textbf{Weight Experiment 3}} & \multicolumn{1}{l|}{\textbf{Weight Experiment 4}} \\ \hline  
\multirow{4}{*}{\shortstack[l]{WEAR\\ (Our approach)}} & Linear Regression & 251.35  & 262.25  & 0.027 & 4.38                                              \\
 & Randon Florest       & \textbf{43.07}                        & 170.93                       & 1.24                                              & 112.75                                            \\
 & Regression Tree             & 60.65                        & \textbf{126.80}                       & 8.53                                              & 20.70                                             \\
 & Lasso          & 251.30                       & 261.95                       & 0.02                                              & \textbf{4.30}                                              \\ \hline
Raykar Algorithm & -- & 256.40                       & 257.77                       & \textbf{0.01}                                              & 6.02                  \\ \hline
\end{tabular}
\end{adjustbox}
\end{table}

\subsection{Real Data }

Next, we investigate the performance of our approach on three datasets: 
\begin{itemize}
    \item \textbf{Dataset 1}: Physio-chemical properties of the tertiary structure of proteins\footnote{\url{https://archive.ics.uci.edu/ml/datasets/Physicochemical+Properties+of+Protein+Tertiary+Structure}} (45730 sample points and 10
variables)
\item \textbf{Dataset 2}:
Tetouan energy consumption\footnote{\url{https://archive.ics.uci.edu/ml/datasets/Power+consumption+of+Tetouan+city}} (48153 sample points and 9 variables)
\item \textbf{Dataset 3}: Prediction of credit card default\footnote{\url{https://archive.ics.uci.edu/ml/datasets/default+of+credit+card+clients}} (52416 sample points and 9 variables)
\end{itemize}

We simulated experts using the true labels in the same way as we do in Section \ref{sec:experiments_simulated}. The variances of the experts were taken to be  (1, 4, 25, 225)   for dataset 1, (1, 9, 64, 40000) for data set 2,  and (1, 9, 64, 40000) for dataset 3.
The dataset was divided into three portions: 70\% for training, 10\% for validation, and 20\% for testing.

The results presented in Table \ref{tab:real} demonstrate that WEAR yields superior performance compared to other methods that do not use $Y$, except for the third dataset, where it matches \citep{R}. Notably, our approach achieves predictive performance that is equivalent to using real labels, which are typically unavailable in practical settings. In contrast, using the arithmetic mean to approximate $Y$ consistently falls short in terms of predictive performance.
Table \ref{tab:weights_real} confirms this by showing that our approach gives better estimates of the variances of each expert in all settings.

\begin{table}[H]
\caption{Estimated Mean Squared Error (MSE) results and corresponding standard errors (in parentheses) for real datasets. Our approach demonstrates comparable performance to methods relying on the gold standard label Y, which is typically unavailable in practical scenarios.}
\label{tab:real}
\begin{adjustbox}{width=\columnwidth,center}
\begin{tabular}{l|l|c|c|c}
\hline
\textbf{Framework}                            & \textbf{Model}                            & \textbf{Dataset 1} & \textbf{Dataset 2} & \textbf{Dataset 3} \\ \hline
  \multirow{4}{*}{\shortstack[l]{WEAR\\ (Our approach)}} & Linear Regression
    & 26.72 (0.34)    & 40688854 (471556.9)    & \textbf{62.99 (1.3)  }   \\
& Random Forest             &  \textbf{12.57 (0.25)}    &  \textbf{26096718 (418018.5)}    & \textbf{62.67 (1.3)}     \\
& Regression Tree                     & 29.56 (0.39)    & 40923181 (495135.2)    & \textbf{63.63 (1.4)}     \\
& Lasso                   & 26.71 (0.34)    & 40692185 (471816.2)                    & \textbf{62.97 (1.3) }    \\ \hline
  Raykar Algorithm  & --                            & 26.72 (0.34)    & 40688854 (471556.9)    & \textbf{62.99 (1.3) }    \\  \hline
  \multirow{4}{*}{Arithmetic Mean}  &Linear Regression  & 26.72 (0.34)    & 1096639436 (4661793.9) & 72.03 (1.8)     \\
& Random Forest              &  13.81 (0.25)    & 1096638824 (4661764.3) & 143.99 (3.1)    \\
& Regression Tree                  & 30.45 (0.39)    & 1096639439 (4661795.0) & 87.39 (1.6)     \\ & Lasso                   & 26.72 (0.34)    & 1096639439 (4661794.0) &  65.62 (1.4)    \\ \hline
  \multirow{4}{*}{
  \shortstack[l]{
  Real $Y$\\ (Gold standard)}} &Linear Regression               & 26.72 (0.34)    & 40688975 (471544.0)    & \textbf{62.98 (1.3) }    \\
& Random Forest                       &  \textbf{12.48 (0.25)}    & \textbf{25984668 (418557.2)}    & \textbf{62.84 (1.3) }    \\
& Regression Tree                            & 29.20 (0.39)    & 40922941 (495146.1)    &\textbf{ 63.63 (1.4)  }   \\
& Lasso                          & 26.72 (0.34)    & 40692151 (471745.3)    &  \textbf{62.94 (1.3)}    \\ \hline
\end{tabular}
\end{adjustbox}
\end{table}

\begin{table}[H]
\caption{Average absolute deviation between each expert's estimated variance and its true value on the real datasets. Our approach provides a more accurate estimate of the experts' precision.}
\label{tab:weights_real}
\begin{adjustbox}{width=\columnwidth,center}
\begin{tabular}{l|l|c|c|c}
\hline
\textbf{Framework}                            & \textbf{Model}                      & \textbf{Weigth Data 1} & \textbf{Weigth Data 2} & \multicolumn{1}{l}{\textbf{ Weigth Data 3}} \\ \hline
  \multirow{4}{*}{
\shortstack[l]{WEAR\\ (Our approach)}}
  & Linear Regression & 24.68                        & 40598542.25                  & 146.14                                           \\
  & Random Florest       & \textbf{16.49}            & \textbf{25318003}                     & 264.77                                           \\
  & Regression Tree             & 28.80                        & 40915427.5                   & \textbf{61.05}                                            \\
  & Lasso             & 24.66                        & 40599886.5                   & 65.02                                            \\ \hline
  Raykar Algorithm  & --                     & 26.17                        & 40175529                     & 318.57                                           \\ \hline
\end{tabular}
\end{adjustbox}
\end{table}

\section{Final remarks}
\label{sec:final}

In conclusion, we have proposed a simple, fast, and powerful method for training in a regression context while taking into account the diversity between labellers. Our approach, named WEAR (Weighted Expertise-based Average Regression), has two steps: estimating the expertise of each labeller by estimating its variability and creating a weighted average of $Y$ using the learned weights. Finally, we regress the weighted average on $\mathbf{x}$ to create our prediction model. We have shown through empirical evidence on both simulated and real data that our approach has better performance than competing approaches.

While our method has demonstrated promising results, there are still areas for improvement. For example, our current approach assumes that the experts are unbiased, which may not always be the case in practice. Future work could explore ways to incorporate expert bias into the weighting scheme. Similarly, WEAR currently assumes that the variances are constant in $\mathbf{x}$, but this may be relaxed. 

Overall, we believe that our approach provides a more flexible and effective solution to the challenge of obtaining accurate labels for regression models with noisy data.  

\section*{Acknowledgement}

This study was financed in part by the Coordenação de Aperfeiçoamento de Pessoal de Nível Superior - Brasil (CAPES) - Finance Code 001.
 
Rafael Izbicki is grateful for the financial support of FAPESP (grant 2019/11321-9) and CNPq (grants 309607/2020-5 and 422705/2021-7).

\section{Appendices}

\appendix

\section{Proofs}

\begin{proof}[Proof of Theorem \ref{thm:main}]

Notice that, under the assumptions of the theorem,
\begin{align*}
h(w_1,\ldots,w_J):&=\E\left[\left(\sum_{j=1}^J  w_j Y_j - Y\right)^2  |\mathbf{x}\right]\\
&= 
\E\left[\left(\sum_{j=1}^J  w_j Y_j - \E[Y|\mathbf{x}]\right)^2  |\mathbf{x}\right]+\E\left[(Y-\E[Y|\mathbf{x}])^2|\mathbf{x}\right]\\
&=\V\left[\sum_{j=1}^J  w_j Y_j|\mathbf{x}\right]+ \V\left[Y|\mathbf{x}\right]    \\
&=\sum_{j=1}^J  w^2_j \V\left[Y_j|\mathbf{x}\right]+ \V\left[Y|\mathbf{x}\right]   
\end{align*}
Now, let
$g(w_1,\ldots,w_{J-1})=h(w_1,\ldots,w_{J-1},1-\sum_{j=1}^{J-1}w_j)$.
It follows that , for every $i \in \{1,\ldots,J-1\}$,
$$\frac{\partial g(w_1,\ldots,w_{J-1})}{\partial w_i}=2w_i\V[Y_i|\mathbf{x}]-2\left(1-\sum_{j=1}^{J-1}w_j\right)\V[Y_J|\mathbf{x}],$$
and therefore the optimal weights satisfy
$$\frac{w_j}{w_1}=\frac{\V[Y_1\mid \mathbf{X} = \mathbf{x}]}{\V[Y_j\mid \mathbf{X} = \mathbf{x}]}.$$
The result follows from the restriction that $\sum_{j=1}^J w_j=1$.
\end{proof}

\begin{proof}[Proof of Theorem \ref{thm:consistent}]
    From the law of large numbers and the fact that
    $$\E\left[\left(Y'_{k,i}- r_i(\mathbf{x}'_k) \right)^2\right]=\V[Y_i|\mathbf{x}],$$
   it follows that $R_i$ is a consistent estimator of $\V[Y_i\mid \mathbf{X} = \mathbf{x}]$. The corollary follows from the properties of convergence in probability.

\end{proof}
\end{document}